\documentclass[11pt]{article}

\PassOptionsToPackage{table}{xcolor}
\usepackage[preprint]{acl}

\usepackage{times}
\usepackage{latexsym}
\usepackage[T1]{fontenc}
\usepackage[utf8]{inputenc}
\usepackage{microtype}
\IfFileExists{inconsolata.sty}{\usepackage{inconsolata}}{}
\usepackage{graphicx}

\usepackage{amsmath}
\usepackage{amssymb}
\usepackage{booktabs}
\usepackage{multirow}
\usepackage{array}
\usepackage{xcolor}
\usepackage{xspace}
\usepackage{enumitem}

\newcommand{\method}{FlashEvolve\xspace}
\newcommand{\aidetwo}{AIDE\textsuperscript{2}\xspace}

\definecolor{bestbg}{RGB}{190,215,240}
\definecolor{secondbg}{RGB}{226,237,249}
\newcommand{\best}{\cellcolor{bestbg}}
\newcommand{\second}{\cellcolor{secondbg}}
\newcommand{\sd}[1]{\,{\scriptsize\color{black!55}$\pm$#1}}
\newcommand{\capgap}{\vspace{-8pt}}
\newcommand{\capgapwide}{\vspace{-12pt}}
\makeatletter\@secpenalty=0\makeatother
\newcommand{\better}{\succ}
\DeclareMathOperator*{\argmax}{arg\,max}

\title{Is Evolution Worth Its Tokens? \\
A Token-Fair Study of LLM Program Search}

\author{
  \textbf{Wujiang Xu}\textsuperscript{1} \quad
  \textbf{Jiaojiao Han}\textsuperscript{2} \quad
  \textbf{Minghao Guo}\textsuperscript{1} \quad
  \textbf{Haolong Jia}\textsuperscript{1} \quad \\
  \textbf{Zhuoran Yin}\textsuperscript{3} \quad 
  \textbf{Mingyu Jin}\textsuperscript{4} \quad
  \textbf{Julian McAuley}\textsuperscript{4} \quad
  \textbf{Dimitris N. Metaxas}\textsuperscript{1} \\
  \textsuperscript{1}Rutgers University \textsuperscript{2}Independent Researcher \\ \textsuperscript{3}Carnegie Mellon University 
  \textsuperscript{4}UC San Diego 
}

\begin{document}
\raggedbottom
\maketitle

\begin{abstract}
Large language models (LLMs) now search for programs that solve optimization problems by writing them, testing them
on public cases and revising them.
Searches spend one token budget in different ways.
Archive-based evolution carries archived programs in every prompt, best-of-N never revises a program, and refinement
revises one program at a time on its public feedback.
We compare these allocations under one token ledger per problem that charges every call and prices each tested
program in tokens.
Our probe is \textbf{\method}, which spends the budget on short refinement attempts that build on the best program,
stop on a stall and restart, by rules that need no model call.
In a pre-registered batch on 40 ALE-Bench problems with Claude Sonnet 4.6, the official five public cases and the
benchmark's own scoring rules, \method scores 140 points above our port of ShinkaEvolve and 116 above best-of-N at
600k tokens.
Refinement from the statement alone scores 158 and 133 points above them, and starting attempts from the best program
adds no separated gain.
Both refinement searches score about 200 points above OpenEvolve.
Pooled over ALE-Bench, CO-Bench and ten contests held after the model's knowledge cutoff, the leads over the port and
best-of-N are 0.75 and 0.65 within-problem standard deviations.
The archive raises the token price of each program, and the port also gains less from each program it tests.
Our code is available at \url{https://github.com/WujiangXu/flashevolve}.
\end{abstract}

\section{Introduction}
\label{sec:intro}

Large language models (LLMs) now evolve programs for optimization problems that have no known exact solution.
FunSearch, AlphaEvolve, OpenEvolve and ShinkaEvolve keep an archive of programs, ask the model to change them, and keep
the changes that test better
\citep{romeraparedes2024mathematical,novikov2025alphaevolve,sharma2025openevolve,lange2025shinkaevolve}.
On the AtCoder Heuristic Contest problems of ALE-Bench and the problems of CO-Bench, such searches and simpler agents
improve programs on public test cases and are scored on hidden ones \citep{imajuku2025alebench,sun2025cobench}.
In AtCoder Heuristic Contest 005, for example, a search writes a C++ program that plans a camera car's patrol route
and revises it on the scores and errors of 50 public maps.
Every call spends tokens, and 600k tokens of Claude Sonnet 4.6 cost about four dollars at list price.
How far such a budget goes depends on how many tokens each tested program takes and on how much each one improves on
the last.
Published systems report their budgets in generations, hours or dollars, and none of those we compare with on the
ALE-Bench Lite subset reports its tokens.

\begin{figure}[t]
  \centering
  \includegraphics[width=\columnwidth]{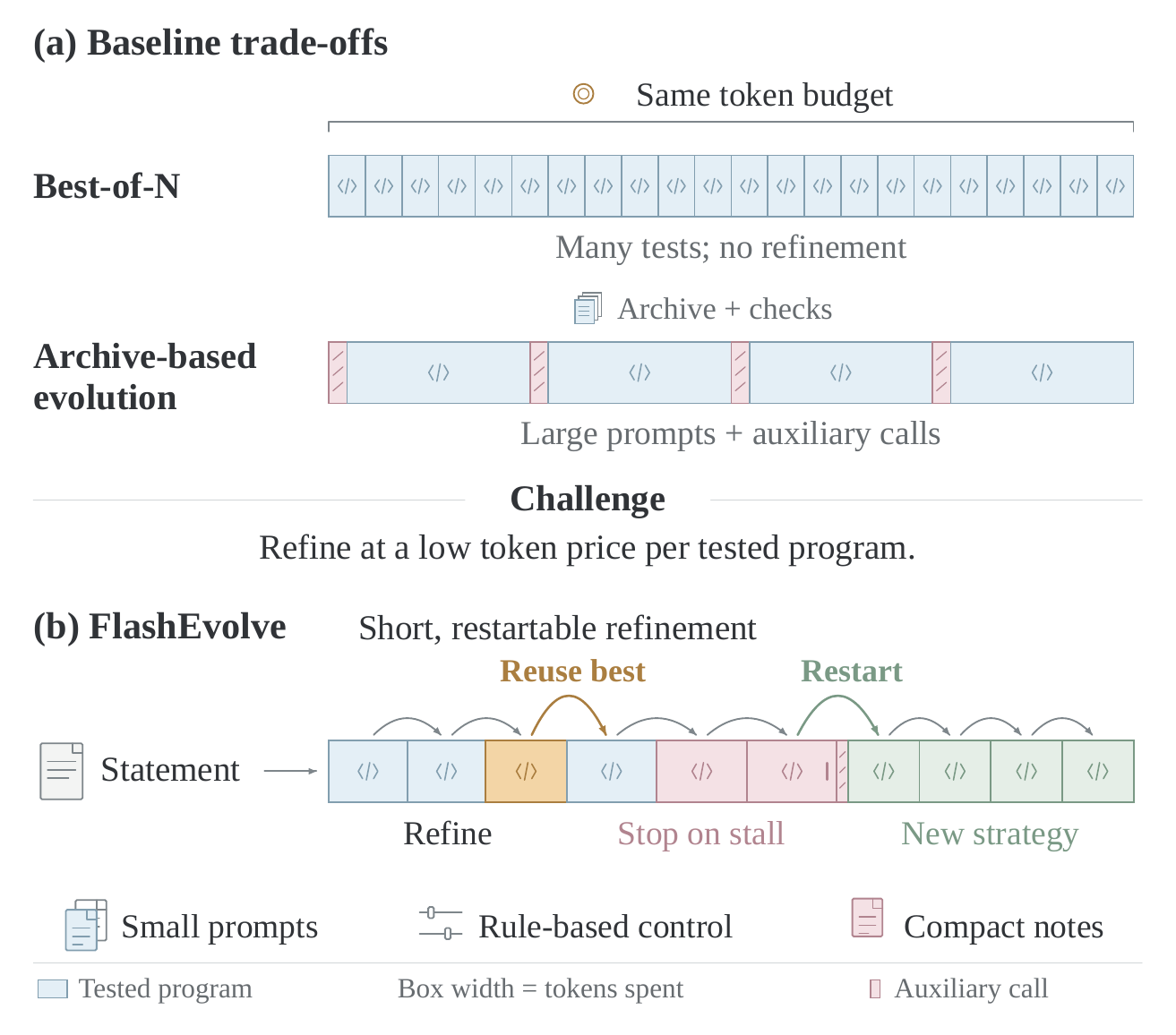}
  \capgap
  \caption{(a) Two ways to spend one token budget, and (b) \method. Each box is a tested program, as wide as the
  tokens that wrote it.}
  \label{fig:intro}
\end{figure}

Existing searches either make each tested program expensive or test programs that are never refined.
ShinkaEvolve writes each program from a prompt that carries a parent program and inspiration programs from its archive
\citep{lange2025shinkaevolve}.
In our port of its default configuration, which starts from the statement and charges every call to the same budget,
inspiration programs fill almost half of each such prompt.
Within 600k tokens the port tests only 26 programs per problem.
Best-of-N makes the opposite trade.
It writes each program in one call and tests 95 programs per problem, yet none of them is ever revised on its
feedback.
The agent that the self-rewriting loop of \aidetwo found shortens what each call carries and returns to the best
program under a new strategy \citep{srikanth2026recursive}.
Its rules came from a self-rewriting run of about eight days whose cost its budget excludes, and it caps the dollars of a run without counting the tokens of each call.

The open question is how one budget should be split between writing new programs and revising existing ones
(Figure~\ref{fig:intro}).
Every archived program that a prompt carries raises the price of the program it writes, and a decision that needs its
own model call spends the budget it is meant to save.
This price can only be measured when every call of a search, including calls that write no program, is counted in one
unit.

We study the split with one token ledger per problem.
It charges every call of a search, including notes, novelty checks and reasoning tokens.
It also stamps each program with the spend at which it returned.
One run then gives the token price of a tested program and the score at every smaller budget.
Our probe is \method, which spends the budget on short refinement attempts whose prompts hold at most two programs and
no archive.
Later attempts start from the best program, stalled ones stop and failed ones restart with a different algorithm, by
rules that read only public results.
Under the same ledger we compare it with sequential refinement, AIDE, best-of-N, independent refinement, AB-MCTS and
two archive-based searches on ALE-Bench, CO-Bench and ten contests held after the model's knowledge cutoff, against
predictions committed before each run.

Refinement carries the result.
With the official five public cases and the benchmark's own scoring rules, \method scores 140 points above our
ShinkaEvolve port and 116 above best-of-N on 40 ALE-Bench problems.
Independent refinement, which starts every attempt from the statement, scores 158 and 133 points above them.
The refinement searches are not separated from each other, and in a pre-registered ablation with five runs per search,
starting attempts from the best program adds no separated gain.
A second archive-based search, OpenEvolve, scores about 200 points below both refinement searches.
The port falls behind because each of its programs takes more tokens and because it gains less from each one, and
given twice the tokens it still scores 108 points below \method.
Without per-case statuses and error messages in the feedback, the model fully repairs a partly failing program in
36\% of next turns against 52\%, although the final score does not separate at 600k tokens.

\section{Related Work}
\label{sec:related}

\subsection{LLM Program Search}

Evolutionary search with language models keeps an archive of programs and asks the model to change them
\citep{romeraparedes2024mathematical,novikov2025alphaevolve,sharma2025openevolve,lange2025shinkaevolve,liu2026evox}.
FunSearch and AlphaEvolve evolve programs for mathematical and algorithmic discovery and build on evolution through
large models and on archives that keep diverse solutions \citep{lehman2022evolution,mouret2015illuminating}.
OpenEvolve, ShinkaEvolve and EvoX make this loop open and add parent sampling, novelty rejection and adaptive
strategies, and the ALE-Bench results of ShinkaEvolve and EvoX start from programs that ALE-Agent wrote.
Agents that start from the statement refine one line of programs or search over many.
ALE-Bench reports sequential refinement, a generic coding agent and ALE-Agent, a search with about one thousand
generations per problem \citep{imajuku2025alebench,wang2025openhands}.
FM Agent reports a cold start without its cost \citep{li2025fm}.
Other systems add memory across runs or a knowledge base without stating where they start
\citep{abaskohi2026sedima,nadafian2026kapso}.
Instead of searching with a fixed agent, \citet{yuksekgonul2026learning} train the model at test time, and AEL
evolves the agent harness in open-ended environments \citep{xu2026ael}.

AIDE and \aidetwo are the closest designs.
AIDE grows a tree of drafts, improvements and fixes and greedily expands the best node \citep{jiang2025aide}.
The self-rewriting loop of \aidetwo found an agent that forks the best program under a new strategy.
This agent also compresses its history into summaries and keeps a memory of failures \citep{srikanth2026recursive}.
\method makes the same two moves, building on the best program and restarting with a different algorithm, by fixed
rules on public results and close to the token price of a short refinement turn.
CORAL runs several autonomous coding agents in separate worktrees that share their best solutions, and it prompts a
change of strategy after repeated failures \citep{qu2026coral}.
It sets a wall-clock budget per task and counts no tokens.
As released, it cannot be stopped at a token count.
We compare published scores only with systems that start from scratch and report a cost, and we rerun \aidetwo's
protocol with our searches (Section~\ref{sec:res-dollar}).

\subsection{Spending a Test-Time Budget}

A test-time budget can buy independent samples, revisions of one program, or experience carried across tasks.
AlphaCode and repeated sampling select the best of many independent programs
\citep{li2022competitionlevel,brown2024large}, and refinement on feedback revises one program with its test results
or a critique \citep{madaan2023selfrefine,chen2024teaching,shinn2023reflexion}.
Self-repair is not always better than new samples at the same budget \citep{olausson2024selfrepair}, and
compute-optimal scaling and hybrid scaling for code divide a budget between parallel samples and sequential revisions
\citep{snell2025scaling,li2025s}.
AB-MCTS makes this split adaptive and decides at every node of a search tree, by Thompson sampling over node scores,
whether to write a new candidate or refine an existing one \citep{inoue2025wider}.
Agents that learn across tasks
\citep{zhao2024expel,ouyang2026reasoningbank,suzgun2026dynamic,xu2026testtime,xu2026mem,wang2025agent,wang2024voyager}
keep their experience as rules, memories or skills, and MemGym tests such memory over long horizons
\citep{xu2026memgym}.
The notes of \method are a small memory of this kind, kept within one problem.

The closest measurement tracks the best solution found at each token budget of an agent and splits a budget across
sessions that start independently \citep{liu2026agents}.
Our ledger records the same curve and adds the token price of a tested program, and \method adds a return to the best
program, patience and restarts to such sessions.

\section{\method}
\label{sec:method}

\begin{figure*}[t]
  \centering
  \includegraphics[width=\textwidth]{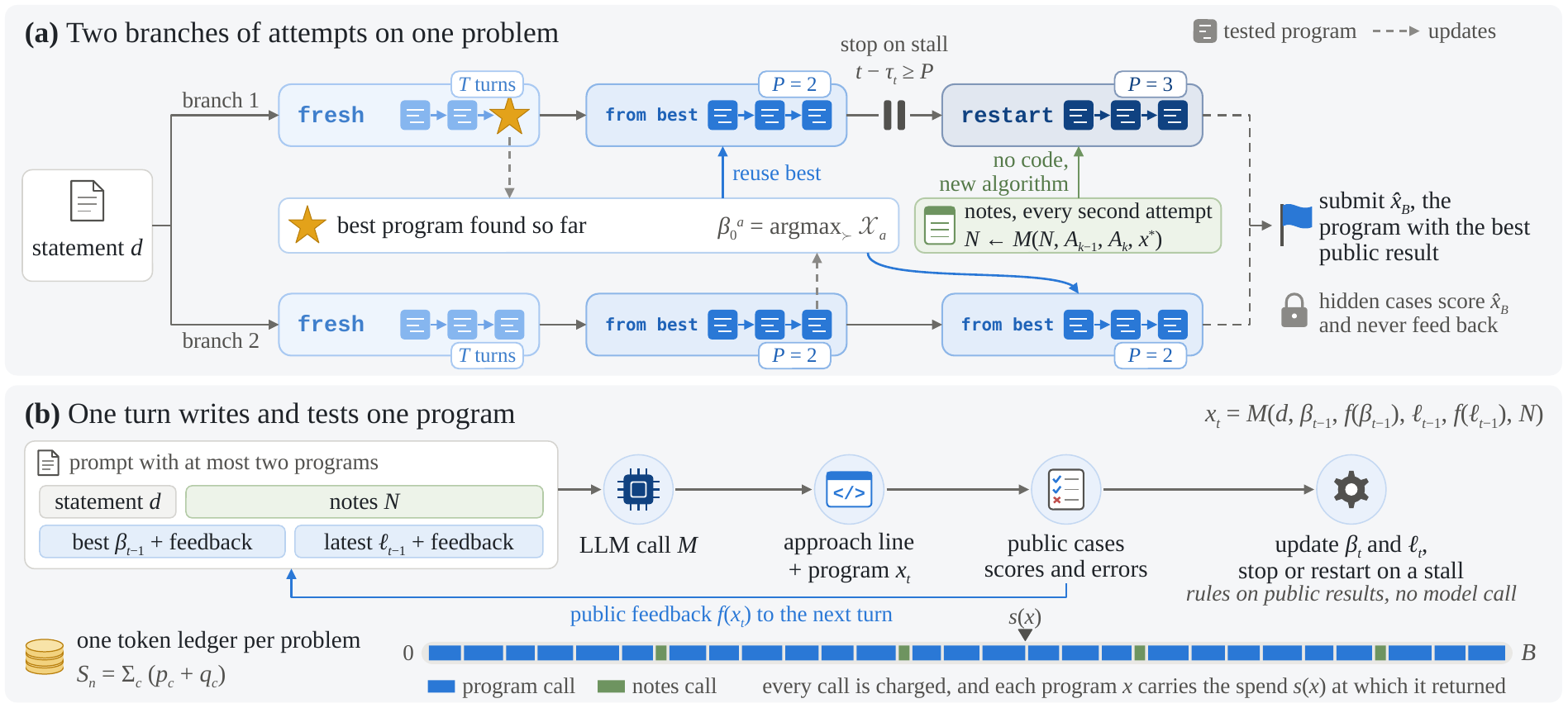}
  \capgapwide
  \caption{\method on one problem. (a) Two branches of short attempts share the best program and the notes. (b) One
turn writes and tests one program, and every call is charged to one token ledger. Symbols follow
Section~\ref{sec:method}.}
  \label{fig:framework}
\end{figure*}

Comparing searches by the token price of a tested program requires counting every call in one unit.
A problem gives the search a statement, public cases whose results the search sees, and hidden cases that only
measure the final program.
\method charges every call to one token ledger per problem.
Let $p_c$ and $q_c$ be the prompt and completion tokens of call $c$, with reasoning tokens counted as completion.
After $n$ calls the ledger holds
\begin{equation}
  S_n = \sum_{c=1}^{n} \left(p_c + q_c\right),
  \label{eq:ledger}
\end{equation}
and call $n+1$ starts only while $S_n$ is below the token budget $B$.
Each program $x$ is stamped with the spend $s(x)$ at which it returned.
Let $v(x)$ count the public cases on which $x$ is valid, let $u(x)$ be its total public score, and let $\sigma$ be
$+1$ when the problem maximizes its score and $-1$ otherwise.
The public result of $x$ is the pair
\begin{equation}
  r(x) = \bigl(v(x),\ \sigma\, u(x)\bigr).
  \label{eq:public}
\end{equation}
We write $x \better y$ when $r(x)$ is larger than $r(y)$ in lexicographic order.
The program submitted at a budget $b \le B$ is the public best among the programs stamped by then,
\begin{equation}
  \hat{x}_b = \argmax_{\better}\ \{\, x : s(x) \le b \,\}.
  \label{eq:submit}
\end{equation}
The token price of a tested program at budget $b$ is the number of tokens spent per program tested within $b$,
\begin{equation}
  \pi_b = \frac{b}{\left|\{\, x : s(x) \le b \,\}\right|}.
  \label{eq:price}
\end{equation}
A search that shows the model more context, or spends calls that write no program, raises $\pi_b$ and tests fewer
programs within the same budget.
Because every program carries its stamp, one run gives the submission, the price and the cost in dollars at list
prices for every smaller budget.

\subsection{Refining a Program on Public Feedback}
\label{sec:turn}

Each call should return a tested program together with the feedback that the next call needs to improve it.
An attempt runs up to $T$ turns, and in each turn the model writes a one-line approach and one program that runs on
the public cases.
Let $M$ be the model, $d$ the statement and $N$ the notes of Section~\ref{sec:notes}.
Within an attempt, let $\beta_{t-1}$ be the best program so far, $\ell_{t-1}$ the latest program and $f(\cdot)$ the
public feedback of a program.
Turn $t$ writes
\begin{equation}
  x_t = M\bigl(d,\ \beta_{t-1},\ f(\beta_{t-1}),\ \ell_{t-1},\ f(\ell_{t-1}),\ N\bigr),
  \label{eq:turn}
\end{equation}
and the attempt keeps $\beta_t = x_t$ if $x_t \better \beta_{t-1}$ and $\beta_t = \beta_{t-1}$ otherwise.
The feedback summarizes the scores, run time and errors of a program on the public cases (Appendix~\ref{app:method}).
The latest program appears only when it did not improve on the best one, and a fresh attempt has no program at its
first turn and writes one from the statement alone.

This turn keeps $\pi_b$ close to the tokens of one call, because a prompt holds at most two programs and no archive
and almost every call returns a program that is then tested.
The turn follows the sequential refinement of ALE-Bench, whose prompt shows the statement, a summary of earlier
attempts and the best and most recent program with their feedback \citep{imajuku2025alebench}.
We show the latest failure beside the best program so that the model can repair a regression instead of repeating it.

\subsection{Building on the Best Program}
\label{sec:stone}

Once the search has found a good program, a new attempt can start from it or from the statement.
\method runs two branches of attempts concurrently, and the first attempt of each branch is fresh.
Every later attempt that is not a restart (Section~\ref{sec:prune}) starts from the best program found so far, and
we call it an attempt from the best program.
Such an attempt re-evaluates this program as its turn zero and shows it to the model with the notes.
Let $\mathcal{X}_a$ be the set of programs written by the attempts that finished before attempt $a$ started.
The attempt starts from
\begin{equation}
  \beta^{a}_{0} = \argmax_{\better}\ \mathcal{X}_a.
  \label{eq:stone}
\end{equation}
A fresh attempt whose program improves on the best program found so far replaces it.
An attempt from the best program starts with a working program and spends its turns on improving it.
It pays for this with a longer prompt in every turn.

\subsection{Stopping and Restarting}
\label{sec:prune}

An attempt that no longer improves still spends one call per turn, and every attempt from the best program inherits
the algorithm of that program.
Let $\tau_t$ be the turn that produced $\beta_t$, zero while $\beta_t = \beta_0$, and let the patience $P$
be two for attempts from the best program and three for restarts.
\method stops an attempt after turn $t$ when
\begin{equation}
  t - \tau_t \ge P.
  \label{eq:patience}
\end{equation}
When an attempt from the best program ends without improving on the best program of the whole search, the next
attempt of its branch
is a restart.
A restart sees the notes and the approach line and public total of the best program but not its code, and its prompt
asks for a substantially different algorithm.
The fresh first attempt of each branch runs all $T$ turns, and restarts never stop before their third turn.
Both rules read only public results, and neither of them needs a call to the model.

\subsection{Carrying Notes across Attempts}
\label{sec:notes}

A restart drops the code of earlier attempts and needs another record of what they tried.
Let $A_k$ hold the approach lines and public results of the $k$-th finished attempt, and let $x^{*}$ be the current
best program with its public total.
After every second finished attempt, one call asks the model for notes of at most 200 words,
\begin{equation}
  N \leftarrow M\bigl(N,\ A_{k-1},\ A_k,\ x^{*}\bigr).
  \label{eq:notes}
\end{equation}
The notes enter every later prompt, and their call is charged to the same ledger.

Each variant of \method is named after the rules of Sections~\ref{sec:prune} and~\ref{sec:notes} that it deletes, as
in \method without restarts.
Early stopping means the patience rule of Equation~\ref{eq:patience} and the restart rule together, and
Appendix~\ref{app:method} gives the prompts and the schedule of attempts.

\begin{table*}[t]
\centering
\small
\setlength{\tabcolsep}{3.8pt}
\renewcommand{\arraystretch}{1.12}
\begin{tabular}{@{}l rrr r rrr rrr@{}}
\toprule
& \multicolumn{3}{c}{Score} & & \multicolumn{3}{c}{\method minus row} &
\multicolumn{3}{c}{Indep.\ refinement minus row} \\
\cmidrule(lr){2-4}\cmidrule(lr){6-8}\cmidrule(l){9-11}
& \multicolumn{2}{c}{Official rules} & \multicolumn{1}{c}{Ours} & & & & & & & \\
\cmidrule(lr){2-3}\cmidrule(lr){4-4}
Search & 5 cases & 50 cases & 50 cases & Prog. & $\Delta$ & W/T/L & Holm $p$ & $\Delta$ & W/T/L & Holm $p$ \\
\midrule
\multicolumn{11}{@{}l}{\emph{Fourth batch, pre-registered. Contrasts with five public cases under official rules.}} \\
\addlinespace[2pt]
Best-of-N$^{\dagger}$ & 1426\sd{13} & 1399\sd{54} & 1431\sd{27} & 96 & $+116^{*}$ & 29/1/10 & 0.014 & $+133^{*}$ & 28/1/11 & 0.010 \\
Port, defaults & 1402\sd{72} & 1412\sd{51} & 1446\sd{22} & 27 & $+140^{*}$ & 27/1/12 & 0.002 & $+158^{*}$ & 32/1/7 & $<$0.001 \\
Sequential refinement & 1458\sd{26} & 1489\sd{111} & 1607\sd{82} & 36 & $+85^{*}$ & 28/0/12 & 0.021 & &  &  \\
AIDE & 1429\sd{19} & 1493\sd{92} & 1577\sd{40} & 47 & $+113$ & 26/1/13 & 0.063 & &  &  \\
Independent refinement & \best 1560\sd{22} & \best 1562\sd{40} & \second 1608\sd{28} & 54 & $-18$ & 21/1/18 & 0.70 & &  &  \\
\method{}$^{\ddagger}$ & \second 1542\sd{88} & \second 1557\sd{89} & \best 1664\sd{39} & 41 & &  & &  & &  \\
\midrule
\multicolumn{11}{@{}l}{\emph{Fifth batch. Contrasts with 50 public cases under our judge.}} \\
\addlinespace[2pt]
OpenEvolve port & & 1362\sd{67} & 1411\sd{36} & 26 & $+194^{*}$ & 35/0/5 & $<$0.001 & $+207^{*}$ & 36/1/3 & $<$0.001 \\
AB-MCTS-A$^{\ddagger}$ & & \second 1562\sd{5} & 1576\sd{24} & 70 & $+29$ & 24/0/16 & 0.62 & $+41^{*}$ & 26/1/13 & 0.011$^{u}$ \\
Independent refinement & & \best 1563\sd{69} & \best 1618\sd{30} & 53 & $-12$ & 16/1/23 & 0.62 & &  &  \\
\method{}$^{\ddagger}$ & & 1531\sd{81} & \second 1605\sd{2} & 41 & &  & &  & &  \\
\bottomrule
\end{tabular}
\vspace{-2pt}
\caption{ALE-Bench, 40 problems, Claude Sonnet 4.6, 600k tokens, mean and standard deviation over runs. The fourth
batch runs each search with five public cases and again with 50, and the fifth adds OpenEvolve and AB-MCTS-A. The
right columns subtract the search of the row from \method or from independent refinement. Shading marks the best two
scores per column and batch, an asterisk an adjusted $p$ below 0.05, and $u$ an unadjusted primary (notes in
Appendix~\ref{app:tablenotes}).}
\label{tab:ale}
\end{table*}

\section{Experimental Setup}
\label{sec:setup}

We evaluate on three benchmarks whose scores are combined only in a standardized pooled reading.
ALE-Bench \citep{imajuku2025alebench} contains 40 AtCoder Heuristic Contest problems.
A search sees either the five public cases of an official Lite session or 50, the protocol of our earlier batches.
We score the submitted program on the hidden cases in AtCoder performance points, which share the scale of published
systems only on the Lite subset of 10 problems.
CO-Bench \citep{sun2025cobench} contains 36 combinatorial optimization problems with at most 40 public and 100 hidden
instances, scored in percent of a reference value.
The held-out contests are the ten AtCoder Heuristic Contests held from April to September 2026, after ALE-Bench and
after the published knowledge cutoff of Claude Sonnet 4.6.
We draw 50 public and 150 hidden cases per contest and score each case in percent of the best result that the
searches of that experiment reached on it.

Every search starts from the statement alone, never sees a hidden case, and gets 600k tokens per problem that also
cover notes, meta-recommendations, novelty checks and reasoning tokens.
Independent refinement runs attempts of six turns from the statement in two branches.
Sequential refinement runs the turn of Section~\ref{sec:turn} as a single chain, and AIDE is our reimplementation of
its greedy tree search \citep{jiang2025aide}.
Best-of-N writes each program in one call.
Our ShinkaEvolve port \citep{lange2025shinkaevolve} runs version 0.0.7 at its defaults, and in earlier batches also
with the published method's meta-recommendation and novelty-check calls.
Our OpenEvolve port \citep{sharma2025openevolve} runs version 0.4.1 at its defaults through an endpoint that charges the
ledger, and AB-MCTS-A runs through the TreeQuest library at its defaults on our refinement turn \citep{inoue2025wider}.
Appendix~\ref{app:port} lists the deviations of each port.

All main experiments use Claude Sonnet 4.6 with three runs per search, and we compare searches only within a batch.
The 600k ALE-Bench runs ran in five batches.
The first three, with 50 public cases, are those of the submitted version (Appendix~\ref{app:earlier}).
The fourth batch is pre-registered and runs the six searches of Table~\ref{tab:ale} twice in one launch, once with
five public cases and once with 50.
With 50 cases, independent refinement and attempts from the best program without notes ran five times.
The fifth adds the OpenEvolve port and AB-MCTS-A at 50 cases.
Further pre-registered batches run the six searches on CO-Bench and the held-out contests, five searches with GPT-5.2
and with Gemini 3 Flash, reduced feedback channels, and the protocol of \aidetwo with Gemini 3 Flash.
We score with the official rules, our reimplementation of the benchmark's own judge, which gives no time slack and,
outside the long contests, scores zero for a program that fails any hidden case.
Our judge allows 1.1 times the time limit and sums the accepted hidden cases, and we name it where it is used.

We average each problem over runs and compare two searches over problems, with W/T/L counting the problems won, tied
or lost.
Differences are in points computed before rounding, tested with the two-sided Wilcoxon signed-rank test
\citep{wilcoxon1945individual} and given 95\% intervals by a hierarchical bootstrap over problems and runs
\citep{efron1993introduction}.
Two searches are separated when $p$ is below 0.05, Holm-adjusted \citep{holm1979simple} within a pre-registered
family where stated.
A pooled reading divides each problem's difference by the standard deviation of its runs and resamples batches and
then problems.
We committed the predictions of every experiment before running it, and Appendix~\ref{app:predictions} lists them with
their outcomes.
\method was developed on ALE-Bench and frozen before CO-Bench and the held-out contests, with attempts of six turns,
two branches, and a temperature of 1.0 for programs and 0.3 for notes.

\begin{table}[t]
\centering
\footnotesize
\setlength{\tabcolsep}{3.2pt}
\begin{tabular}{@{}lrrr@{}}
\toprule
System & Cost & Off. & Ours \\
\midrule
\multicolumn{4}{@{}l}{\emph{Published systems, as reported}} \\
OpenHands, o4-mini-high & \$2.26 & 859 & \\
OpenHands, Gemini 2.5 Pro & \$3.30 & 903 & \\
Seq.\ ref., o4-mini-high & \$7.17$^{a}$ & 1411 & \\
Seq.\ ref., Gemini 2.5 Pro & \$11.13$^{a}$ & 1198 & \\
ALE-Agent, Gemini 2.5 Pro & & & \\
\quad base & \$7.64 & 1167 & \\
\quad + domain & \$11.12 & 1264 & \\
\quad full & \$100.33 & 1879 & \\
AIDE0, Gemini 3 Flash & $\le$\$5$^{b}$ & 1536 & \\
AIDE\textsubscript{human}, Gemini 3 Flash & $\le$\$5$^{b}$ & 1511 & \\
AIDE85, Gemini 3 Flash & $\le$\$5$^{b,c}$ & 1790 & \\
AIDE0, AIDE85, stronger models & $\le$\$20$^{e}$ & \multicolumn{2}{@{}l}{1796 to 2121} \\
\midrule
\multicolumn{4}{@{}l}{\emph{Ours, Claude Sonnet 4.6, first batch}} \\
\method & \$4.42$^{d}$ & \second 1504 & \best 1620 \\
\quad w/o early stopping & \$4.27$^{d}$ & \best 1534 & \second 1597 \\
ShinkaEvolve port & \$3.56$^{d}$ & 1410 & 1406 \\
\midrule
\multicolumn{4}{@{}l}{\emph{Ours, Claude Sonnet 4.6, second batch}} \\
\method & \$4.40$^{d}$ & \best 1442 & \best 1564 \\
\quad w/o early stopping & \$4.27$^{d}$ & 1391 & \second 1538 \\
Sequential refinement & \$4.15$^{d}$ & 1380 & 1504 \\
AIDE & \$4.53$^{d}$ & \second 1441 & 1519 \\
Best-of-N & \$6.11$^{d}$ & 1381 & 1383 \\
\midrule
\multicolumn{4}{@{}l}{\emph{Ours, Gemini 3 Flash, protocol of AIDE0, 10 runs}} \\
\method & $\le$\$5$^{f}$ & 1480 & 1622 \\
Independent refinement & $\le$\$5$^{f}$ & \best 1566 & \best 1665 \\
AIDE & $\le$\$5$^{f}$ & \second 1493 & \second 1595 \\
\bottomrule
\end{tabular}
\capgap
\caption{ALE-Bench Lite, 10 problems. Off.\ is official rules and Ours our judge. The last block caps each problem and
run at five dollars and shows five public cases (notes in Appendix~\ref{app:tablenotes}).}
\label{tab:lite}
\end{table}

\section{Results}
\label{sec:results}

\begin{figure*}[t]
  \centering
  \includegraphics[width=\textwidth, trim=0 5 0 0, clip]{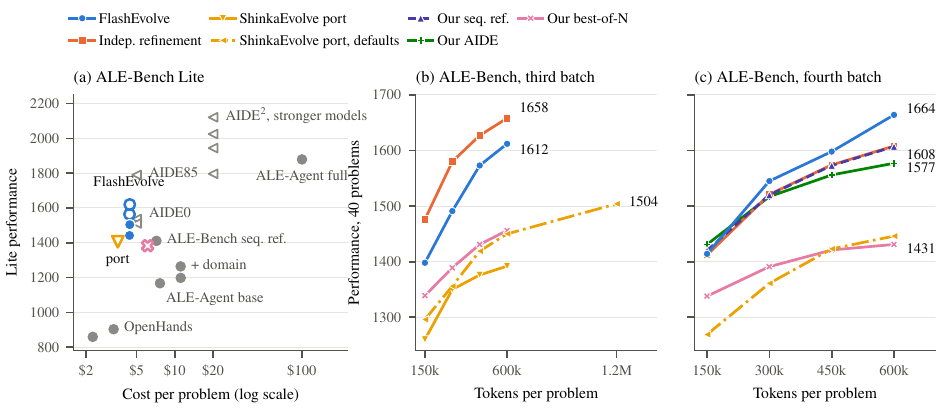}
  \capgapwide
  \caption{(a) Lite performance at reported cost or our list price. Triangles mark cost caps, and our searches with
  Claude show official rules as filled and our judge as open markers. (b, c) 40 problems, 50 public cases, our judge,
  with the port at its defaults run to 1.2M tokens in (b).}
  \label{fig:cost}
\end{figure*}

\begin{figure*}[t]
  \centering
  \includegraphics[width=\textwidth, trim=0 8 0 7, clip]{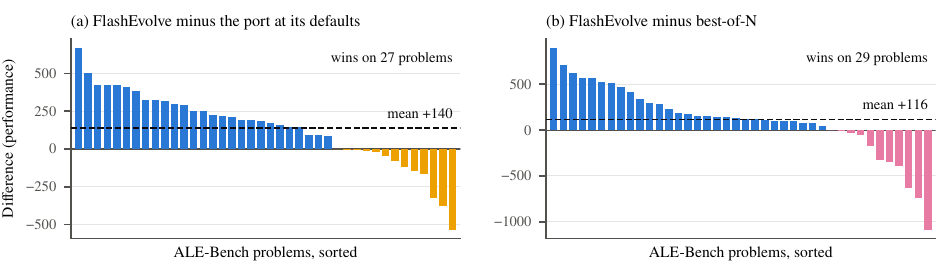}
  \capgapwide
  \caption{\method minus our port at ShinkaEvolve's defaults (a) and minus best-of-N (b) on each of the 40 ALE-Bench
  problems, sorted, in the pre-registered fourth batch with five public cases under official rules.}
  \label{fig:perproblem}
\end{figure*}

\begin{table}[t]
\centering
\footnotesize
\setlength{\tabcolsep}{2.6pt}
\begin{tabular}{@{}lrrrr@{}}
\toprule
& & & \multicolumn{2}{c}{ALE 40, ours} \\
\cmidrule(lr){4-5}
Search & CO-Bench & Held-out & GPT-5.2 & Gemini \\
\midrule
Best-of-N & 84.8\,{\scriptsize$\pm$0.4} & \second 75.2\,{\scriptsize$\pm$0.7} & 1578 & 1843 \\
Port, defaults & 85.4\,{\scriptsize$\pm$1.2} & 62.1\,{\scriptsize$\pm$1.7} & n.r. & n.r. \\
Seq.\ refinement & 86.6\,{\scriptsize$\pm$0.4} & 65.2\,{\scriptsize$\pm$2.8} & 1523 & 1834 \\
AIDE & \second 87.6\,{\scriptsize$\pm$2.0} & 75.1\,{\scriptsize$\pm$10.5} & 1480 & \best 1876 \\
Indep.\ refinement & 87.4\,{\scriptsize$\pm$0.4} & 73.2\,{\scriptsize$\pm$3.7} & \best 1656 & 1868 \\
\method & \best 89.4\,{\scriptsize$\pm$0.8} & \best 77.2\,{\scriptsize$\pm$2.0} & \second 1635 & \second 1874 \\
\bottomrule
\end{tabular}
\capgap
\caption{CO-Bench (36 problems, percent of the reference) and the ten held-out contests (percent of the best of these
six searches per case), Claude Sonnet 4.6, and ALE-Bench with two more models under our judge, at 600k tokens per
problem (notes in Appendix~\ref{app:tablenotes}).}
\label{tab:beyond}
\end{table}

\subsection{Searches at One Token Budget}
\label{sec:res-token}

Every search runs on the 40 ALE-Bench problems with Claude Sonnet 4.6 at 600k tokens, and Table~\ref{tab:ale} gives
the pre-registered fourth batch.
With the official five public cases and official rules, \method leads our port in ShinkaEvolve's default
configuration by 140 points and best-of-N by 116, the two co-primaries of the batch (Figure~\ref{fig:perproblem}).
Independent refinement leads them by 158 and 133 points.
Best-of-N tests 96 programs per problem and revises none of them, and the port tests 27.
Testing many programs helps only when they are refined, and the archive in every prompt of the port leaves it fewer
programs to test.

The refinement searches are close to one another.
Under official rules with five cases \method leads sequential refinement by 85 points and AIDE by 113, the first
separated after Holm correction, and it scores 18 points below independent refinement without separation.
At 50 cases none of the four refinement searches is separated from another under either judge.
AB-MCTS-A, which chooses at every node between writing and refining, scores 41 points below independent refinement
under our judge ($p=0.011$) and 1 point below under official rules.
It spent about 40\% of its nodes on fresh programs at the root, with a mean depth of 0.8.
The OpenEvolve port tests 26 programs per problem and scores 207 points below independent refinement and 194 below
\method under our judge, on 36 and 35 of the 40 problems.

The protocol moves every search, and the leads stay.
At 50 public cases the two co-primaries are 146 and 158 points under official rules and 218 and 233 under our judge.
Re-scoring every stored submission shows that the time allowance of our judge is worth 9 to 25 points and its scoring
rule 37 to 56, and the two effects are additive (Appendix~\ref{app:judge}).
With one public case the search of \method still leads best-of-N by 135 points under our judge ($p=0.018$), and with
the five smallest inputs by 186 ($p=0.004$).
Choosing the submission with five cases instead of 50 costs each search of the third batch 10 to 35 points, and the
leads of \method in that batch hold under every subset.
The earlier batches agree, with leads of 162 points over the port and 156 over best-of-N in the pre-registered third
batch (Appendix~\ref{app:earlier}).

\subsection{Other Benchmarks and Models}
\label{sec:res-beyond}

To check whether the result depends on the development benchmark or on Claude, we froze every search and ran the six
of Table~\ref{tab:ale} on CO-Bench and the ten held-out contests (Table~\ref{tab:beyond}).
\method leads the port on both, by 4.0 points of the CO-Bench reference (25/2/9, Holm $p=0.0009$) and by 15.4 points of
the best case on the contests (9/0/1, Holm $p=0.020$).
Its leads over best-of-N, 4.6 and 2.3 points, are not separated on either benchmark.
We pre-registered a pooled reading over CO-Bench, the contests and the 50-case ALE-Bench runs, 84 problems in all.
There \method leads the port by 0.75 within-problem standard deviations [0.57, 0.98] and best-of-N by 0.65 [0.31,
0.95], each benchmark on its own is positive, and \method is not separated from independent refinement.
Without the ten ALE-Bench problems used during development the lead over best-of-N is 0.54 [0.28, 0.85].

With GPT-5.2 and Gemini 3 Flash, three runs of five searches each, the searches lie within 175 points on GPT-5.2 and
42 on Gemini 3 Flash.
Independent refinement leads best-of-N by 78 points on GPT-5.2 (Holm $p=0.014$), and pooled over the three model
families \method leads best-of-N by 0.43 within-problem standard deviations [0.10, 0.80].

\subsection{Cost against Published Systems}
\label{sec:res-dollar}

On the Lite subset our scores share the scale of published systems (Table~\ref{tab:lite}, Figure~\ref{fig:cost}a).
For about four dollars per problem, every search we ran with Claude Sonnet 4.6, best-of-N and the port included,
scores above OpenHands, the two cheaper settings of ALE-Agent and the sequential refinement of ALE-Bench with Gemini
2.5 Pro.
We attribute this placement to the model and our protocol.

We reran the protocol of \aidetwo with three of our searches, Gemini 3 Flash, five dollars per problem and run at
list prices, five public cases, official rules and ten runs.
Our AIDE scores 1493 against 1536 for the AIDE0 of \aidetwo, a gap of 43 points between the two harnesses.
\method scores 1480 and independent refinement 1566, with standard errors of 61 and 51, and no difference among our
three searches is separated after Holm correction.
AIDE85 reports 1790, and the 85 rounds in which a stronger model rewrote its inner agent lie outside its cap.
The systems that remain above our searches either spend more per problem or improve the agent itself, and with
1.8M tokens \method still stays below AIDE85 and ALE-Agent full (Appendix~\ref{app:earlier}).

\section{Analysis}
\label{sec:analysis}

\subsection{Price and Gain of the Port}
\label{sec:ana-port}

The port falls behind both because each tested program takes more tokens and because it gains less from each program.
In the pre-registered third batch the port at its defaults spends 480k tokens on its first 20 programs against 298k
for \method and tests 26 programs within 600k against 39 (Figure~\ref{fig:price} in Appendix~\ref{app:price}).
With 1.2M tokens it tests 47 programs, more than the 39 of \method at 600k, and still scores 108 points lower under
our judge, separated in a pre-registered test (Figure~\ref{fig:cost}b).

The calls of the port that write no program take a quarter of its tokens in our configuration, and the default
configuration, which drops them, scores 58 points higher without being separated (Appendix~\ref{app:price}).

\subsection{Ablations of the Rules}
\label{sec:ana-parts}

The fourth batch tests the start from the best program directly, with the same budget, branches and turns and five
runs each.
Attempts from the best program without notes lead attempts from the statement by 41 points under our judge (22/1/17,
$p=0.315$), inside the 47 points that five runs resolve.
Pooled over 83 problem and batch pairs the difference is 0.14 within-problem standard deviations, with an interval
that includes zero.
The advantage of \method over best-of-N and the port is an advantage of refinement itself.

Patience and restarts do not separate \method from its variants (Appendix~\ref{app:parts}).
After a restart prompt asks for a different algorithm, the next program changes algorithm family in 86\% of restarts,
against 66\% of attempts from the best program, in a judge's reading of 47,190 stored turns.
The next attempt implements 39 to 54\% of the recommendations in the notes, and removing the notes does not lower the
score.

\subsection{Language in the Feedback}
\label{sec:ana-language}

In a pre-registered batch we cut the feedback text of independent refinement and \method, which lists per-case
statuses, scores, run times and error messages.
With only the number of valid cases and the total score, the share of partly failing programs that the next turn
makes fully valid falls from 52\% to 36\% for independent refinement and from 57\% to 35\% for \method.
Dropping only the error messages lowers the first share to 41\%.
The submitted score falls by 13 to 42 points with every reduction, and none of these drops is separated at 600k
tokens.

\section{Conclusion}
\label{sec:conclusion}

At a fixed token budget, refinement on public feedback is the allocation that pays.
With the official five public cases and official rules, \method and independent refinement score 116 to 158 points
above best-of-N and our ShinkaEvolve port on 40 ALE-Bench problems, and both score about 200 points above OpenEvolve.
The lead over the port holds on CO-Bench and on ten contests held after the knowledge cutoff.
The rules that \method adds to refinement show no separated gain, and AB-MCTS-A, which adapts the split between
writing and refining, does not improve on fixed width either.
The next test is a budget at which these rules could pay, for instance on problems where fresh starts rarely find a
valid program.

\section*{Limitations}

The lead over best-of-N is separated on ALE-Bench only.
On CO-Bench it is 4.6 points and on the held-out contests 2.3, neither separated, and with GPT-5.2 and Gemini 3 Flash
it is not separated on either model alone.
Under official rules with five cases it rests on the ten ALE-Bench problems used during development, 260 points there
against 67 on the other 30 ($p=0.17$), although the pooled reading without those ten stays separated.
Under the protocol of \aidetwo, \method scores 1480, below the reported AIDE0 at 1536 and AIDE85 at 1790 and below our
own independent refinement, and our five-dollar cap counts LLM tokens without the execution that \aidetwo includes.
AB-MCTS-A is 41 points below independent refinement under our judge and level with it under official rules, and its
node score, the rank of a program among the valid programs of its run, is our choice.
Our archive-based baselines are our own ports of ShinkaEvolve and OpenEvolve with the deviations of
Appendix~\ref{app:port}, run at 600k tokens per problem, and the ShinkaEvolve port at its defaults still gains 54
points from 600k to 1.2M tokens.
Reduced feedback lowers repair rates without a separated change in score, and approach lines name the main method of
their program in 75\% of a judge's sample.
The restart prompt quotes these lines and carries their errors.
\method scored zero on toyota2023summer-final in 9 of its 18 runs over the five batches.
In all nine, every attempt from the best program started from a program that passed no public case.
Building on a program only after it passes a case gave a valid program in two of three runs
(Appendix~\ref{app:toyota}).
Three runs per search resolve about 61 points on 40 problems, 2.5 points on CO-Bench and 12 points on the held-out
contests, and smaller differences stay unresolved.
The training data of the models may reach past the dates of the held-out contests.

\bibliography{references}
\clearpage
\twocolumn
\tableofcontents

\clearpage
\appendix

\section{\method in Detail}
\label{app:method}

Every turn sends a system message that casts the model as a programmer for AtCoder Heuristic Contest problems.
The statement follows, and then the notes, the best program of the attempt and the latest program when they exist.
The model must answer with one line that summarizes its algorithm and one complete program.
The public feedback of a program lists status counts, the mean and total score, the scores of the first 20 cases, the
slowest time and up to three error messages.
Figure~\ref{fig:prompts} quotes the parts of the prompt that change between turns.

\begin{figure*}[t]
\centering
\begin{minipage}{0.92\textwidth}
\footnotesize
\begin{verbatim}
first turn of a fresh attempt:
  Write a complete, strong first solution.
later turns:
  # Current best program (turn t; approach: ...)   <code>  <public feedback>
  # Most recent attempt (turn t; approach: ...), not better than best
                                                    <code>  <public feedback>
every turn:
  Output format (strict): first one line `APPROACH: <one-sentence summary
  of the algorithm>`, then one complete program.
restart:
  ## Restart
  The current best program uses: <approach> (public total <total>).
  It has stopped improving. Write a new solution based on a
  substantially DIFFERENT algorithmic idea.
notes request (every second finished attempt):
  task_notes: <=200 words for whoever continues THIS problem ...
\end{verbatim}
\end{minipage}
\capgap
\caption{The parts of the \method prompt that change between turns, quoted from our code.}
\label{fig:prompts}
\end{figure*}

Two branches share the ledger of a problem, and each branch runs attempts until the ledger reaches the budget.
The first attempt of a branch is fresh and runs six turns.
A later attempt starts from the best program, unless the previous such attempt of the branch ended without improving
on the best program of the whole search, in which case it is a restart.
An attempt from the best program re-evaluates that program as its turn zero and stops after two turns without
improving on its own best program.
A restart stops after three such turns but never before its third turn.
After every second finished attempt one call rewrites the notes.
The submission at a budget is the program with the best public result among those stamped within it.
Programs run in a container without network access and are compiled with gcc 14 at optimization level two, and a run
of the hidden cases never feeds back into the search.

\section{Predictions and Outcomes}
\label{app:predictions}

We committed the predictions of each experiment before running it, and Table~\ref{tab:predictions} lists those that
concern the searches we report, with their outcomes.
We pre-registered the third batch with its predictions and with rules for this paper that were accepted before
launch.
One rule said that a lead of \method over the port in its default configuration below 180 points would replace the
first batch's 241 in the abstract and introduction, and it fired.
The first two batches' contrasts against the port and best-of-N were not themselves predicted, and neither were the
numbers on published systems, attempt kinds or the problem left out in Table~\ref{tab:ale}.
The third batch's pre-registered floor for its primary contrast was 160 points, after the original instruction
reached the experiment session cut short.
The early-stopping policy of Appendix~\ref{app:learned} was launched after two of its seven gates failed in an offline replay and were removed as launch criteria.

\begin{table*}[t]
\centering
\footnotesize
\setlength{\tabcolsep}{4pt}
\begin{tabular}{@{}p{2.9cm}p{6.4cm}p{6.0cm}@{}}
\toprule
Experiment & Primary and other predictions & Outcome \\
\midrule
Third batch, ALE 40 & \method minus the port at its defaults at least $+160$ with $p<0.05$. \method minus the port in ours at least $+120$. \method above best-of-N. The port at its defaults above the port in ours. Both \method leads under official rules. \method at 600k above the port at its defaults at 1.2M. & $+161.6$, $p$ 2.5e-05, met. $+219.6$, met. $+156$, met. $+58$, not separated. $+143$ and $+236$, met. $+107.8$, met. \\
Second batch, ALE 40 & W/o early stopping minus sequential refinement at least $+30$ with $p<0.05$. Notes add at least 15. W/o early stopping minus AIDE at least $+20$. \method above sequential refinement. W/o early stopping minus best-of-N at least $+100$. & $+26.6$, $p$ 0.221, not met. Notes $-35.3$, not met. $+66.1$, met. $+4.7$, met as stated but reported without a claim. $+157.6$, met. \\
Two more models & Gemini Lite rows within $\pm200$ of AIDE0. & Met. \\
Learned early stopping & Three predictions on reaching earlier searches with fewer tokens. & One of three held. \\
Budgets to 1.8M, Lite & \method minus w/o early stopping at least $+15$ at 1.8M. Every search improves past 600k. & $+70.5$ on the mean, $p$ 0.275, met as stated. Met. \\
Analyses of stored runs & Official rules keep the Lite ordering. & One of three held. \\
\midrule
Fourth batch, ALE 40, five cases & \method minus the port and minus best-of-N above 0 under official rules, Holm over the pair. \method minus independent refinement within $\pm61$. & $+139.8$, Holm 0.002, held. $+115.6$, Holm 0.014, held. $-17.7$, held. \\
Fourth batch, ALE 40, start & Attempts from the best program without notes minus independent refinement, five runs each, primary. & $+41.2$, $p$ 0.315, within $\pm47$, not separated. \\
Fifth batch, ALE 40 & \method and independent refinement minus OpenEvolve above 0, Holm over the pair. AB-MCTS-A minus independent refinement, two-sided. & $+194.4$ and $+206.9$, held. $-41.4$, $p$ 0.011 under our judge, $-0.9$ under official rules. \\
CO-Bench and held-out & \method minus the port and minus best-of-N above 0 on each, Holm over the pair. Both above 0 pooled over three benchmarks. & Port held on both, $+4.0$ and $+15.4$. Best-of-N not separated, $+4.6$ and $+2.3$. Pooled held, 0.75 and 0.65. \\
Two more models, three runs & \method and independent refinement minus best-of-N above 0 on each model. Pooled over three model families. & One of four held, independent refinement on GPT-5.2 at $+78.2$. Pooled held, 0.43. \\
Reduced feedback, ALE 40 & Full feedback minus the valid count and total above 0, for independent refinement and \method. & $+12.8$ and $+31.1$, not separated. \\
Protocol of \aidetwo, Lite & Our AIDE within $\pm100$ of AIDE0. Contrasts among three searches, two-sided. & 1493 against 1536, held. None separated. \\
Re-scoring and re-selection & The scoring rule costs more than the time allowance. Selection with five cases costs under 60 points. & Held on four batches. Held, 10 to 35. \\
Language in stored turns & Restarts change algorithm family more often than attempts from the best program, and four further expectations. & Held, 86\% against 66\%. One further expectation held in part and three failed. \\
\bottomrule
\end{tabular}
\capgap
\caption{Predictions on the searches reported here, committed before each experiment, and their outcomes. The rows
below the rule belong to the revision. Differences are in the unit of each benchmark, and w/o means without.}
\label{tab:predictions}
\end{table*}

\section{Protocol and Judges}
\label{app:judge}

Our harness reimplements ALE-Bench from its released data and differs from the official one in four ways.
Our searches see 50 public cases instead of the five of an official Lite session, and our judge allows 1.1 times the
time limit.
On the short contests it sums the accepted hidden cases where the official judge scores a program zero unless every
case is accepted, and we compile without the AtCoder Library and Boost.
Official rules removes the slack and applies the all-accepted rule, with the long contests of the Lite subset exempt
as derived from our data.

Under official rules the Lite score of \method falls by 116 points in the first batch, where two problems lose 635 and
557 points.
It falls by 122 points in the second batch, where two problems lose 612 and 584 points.
Official rules were scored in a different session from our judge, whose drift between sessions is about five points
per search.
Of its 100 scoring shards, 14 ran on loaded machines, which can only lower scores.

The fourth batch makes the official five public cases and official rules the main protocol, and we measured the
three ways in which our earlier protocol differs.
Re-scoring every stored submission of four batches with the time allowance and the scoring rule switched separately
shows that removing the allowance lowers the mean of each search by 9 to 25 points under the summed rule.
The all-accepted rule lowers them by 37 to 56 points without the allowance, and the interaction of the two is within
25 points on every batch.
Re-choosing each submission of the third batch from its stored programs with the official five of the 50 public
cases costs 10 to 35 points across its five searches.
The leads of \method over the port and over best-of-N stay separated under every subset we tried, including the five
smallest and the five largest inputs.
The rank correlation between public and hidden scores rises from 0.45 with one case to 0.80 with 50 for best-of-N.
Rerunning the search with fewer cases, two runs each, gives \method a lead over best-of-N of 135 points with one case,
205 with ten and 186 with the five smallest inputs, all with $p$ below 0.02 under our judge.

\section{Patience, Restarts and Notes}
\label{app:parts}

\begin{figure}[t]
  \centering
  \includegraphics[width=\columnwidth]{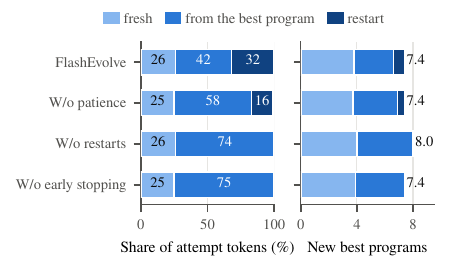}
  \capgap
  \caption{Attempt tokens and new best programs by kind of attempt, second batch, per problem and run.}
  \label{fig:kinds}
\end{figure}

Table~\ref{tab:kinds} gives the attempts behind Figure~\ref{fig:kinds}.
Patience shortens attempts from the best program and makes room for more of them.
In the two-by-two comparison of patience and restarts, patience changes the score by $+16.5$ without restarts and
$+0.6$ with them, and restarts change it by $-22.6$ without patience and $-38.5$ with it.
Every interval covers zero and every Holm-adjusted $p$ value is 1.
Without toyota2023summer-final, on which \method scored zero, the two contrasts that involve \method become $+37.4$
and $-7.4$.

\begin{table}[t]
\centering
\footnotesize
\setlength{\tabcolsep}{2pt}
\begin{tabular}{@{}llrrrr@{}}
\toprule
Search & Kind & Attempts & Turns & Share & Bests \\
\midrule
\method & fresh & 2.0 & 6.0 & 26\% & 3.8 \\
 & from best & 5.7 & 2.8 & 42\% & 2.8 \\
 & restart & 3.0 & 4.7 & 32\% & 0.8 \\
W/o early stopping & fresh & 2.0 & 6.0 & 25\% & 3.9 \\
 & from best & 5.5 & 5.1 & 75\% & 3.5 \\
W/o restarts & fresh & 2.0 & 6.0 & 26\% & 4.0 \\
 & from best & 10.6 & 2.6 & 74\% & 4.0 \\
W/o patience & fresh & 2.0 & 6.0 & 25\% & 3.7 \\
 & from best & 4.2 & 5.2 & 58\% & 3.2 \\
 & restart & 1.5 & 4.9 & 16\% & 0.5 \\
\bottomrule
\end{tabular}
\capgap
\caption{Attempt kinds in the second batch, per problem and run. Turns are per attempt, Share is each kind's part of
the attempt tokens, Bests counts new best programs, and w/o means without.}
\label{tab:kinds}
\end{table}

The notes are short and accurate.
In the first batch, \method writes its first note after a median of 154k tokens and spends 6.1\% of its tokens on
notes.
Of 1,725 checkable numerical claims in its notes, 94\% match a public result logged before the note.
In the second batch \method without early stopping or notes scored 1684, the highest score of that batch, and \method minus
this variant is $-57.3$ points with a Holm-adjusted $p$ of 0.751.

\section{Price of a Tested Program}
\label{app:price}

\begin{figure}[t]
  \centering
  \includegraphics[width=\columnwidth]{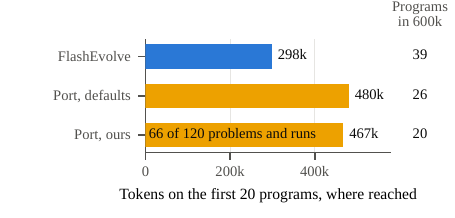}
  \capgap
  \caption{Tokens on the first 20 programs in the pre-registered third batch, where reached, with the programs each
  search tests within 600k tokens.}
  \label{fig:price}
\end{figure}

Table~\ref{tab:price} compares searches at the same number of tested programs instead of the same number of tokens.
At each count, a search submits the program with the best public result among its first programs.
A problem enters the rows on the port only when every first-batch run tested at least that many programs on it.
The port averages 19 programs per problem, and these rows shrink from 40 problems at five programs to 11 at twenty.
We specified them after the main results.
The bottom rows give the programs that each search tests within 600k tokens with each model.
Without the charges of its recommendation and novelty calls, the curve of the port moves by 17 and 19 points at 150k
and 300k tokens.
At larger budgets this uncharged reading runs past the end of the stored runs, whose uncharged spend ends near 457k
tokens.

\begin{table}[t]
\centering
\footnotesize
\setlength{\tabcolsep}{3.6pt}
\begin{tabular}{@{}lrrrr@{}}
\toprule
Programs tested & 5 & 10 & 15 & 20 \\
\midrule
FE $-$ port, ALE 40 & $-37$ & $-13$ & $+52$ & $+154$ \\
W/o e.s.\ $-$ port, ALE 40 & $+4$ & $+68$ & $+134^{*}$ & $+143$ \\
\quad problems covered & 40 & 35 & 26 & 11 \\
\midrule
Programs within 600k & & & W/o e.s. & FE \\
Claude Sonnet 4.6 & & & 38 & 40 \\
GPT-5.2 & & & 37 & 41 \\
Gemini 3 Flash & & & 20 & 21 \\
\bottomrule
\end{tabular}
\capgap
\caption{Paired differences at equal program counts in the first batch, on the problems on which every run reached
the count, and programs tested within 600k tokens. Only the starred interval excludes zero ($p$ 0.032, unadjusted).
The bottom rows cover the 40 ALE-Bench problems, with Claude Sonnet 4.6 in the second batch. FE is \method and W/o
e.s.\ is without early stopping.}
\label{tab:price}
\end{table}

\section{Beyond ALE-Bench and Claude}
\label{app:beyond}

On the held-out contests \method scores 80.8 and \method without early stopping 70.1 percent of the best result per case.
On ahc068 every search is valid on all cases and the score falls with the number of operations.
The submissions of \method without early stopping use about 79,000 operations there and score 6.8, and \method scores 36.8.
\method without early stopping or notes, which also builds on the best program, scores 83.1 in a later batch, and the one run
of \method that found the short solution found it in a fresh attempt.
Choosing the final program by public result loses 0.9 points for \method and 0.4 for \method without early stopping against
an oracle that chooses by the hidden cases, with each case's score capped at 100 percent.
With GPT-5.2 and Gemini 3 Flash on the 40 ALE-Bench problems, \method scores 1652 and 1849, and \method without early
stopping scores 1601 and 1871 in the same batch.

\section{Learned Early Stopping}
\label{app:learned}

\method with learned early stopping replaces the two fixed rules with a policy.
At each boundary it takes the action with the highest expected public gain per token, estimated from priors fitted on
development runs and from the run's own public results.
Two of its seven launch gates failed in an offline replay and were removed as launch criteria, and the version that
ran passes one of them only at equality and fails the other.
It scores 88.0 on CO-Bench and 69.0 on the held-out contests, the lowest of the searches there, and 1677 on the Lite
subset at 1.8M tokens, where it is not separated from \method.

\section{The ShinkaEvolve Port}
\label{app:port}

The port runs ShinkaEvolve 0.0.7 with two islands, an archive of 40 programs, and diff, full and crossover patches, in
two configurations.
The default configuration of 0.0.7 makes no meta-recommendation or novelty-check calls.
Ours switches both on, as the published method describes, and they take a quarter of its tokens.
Both use one model, replace the unavailable embedding model with hashed code embeddings, and write their seed program
in a repair chain inside the same budget, which on one problem took 279k of 618k tokens.
In the default configuration a tested program takes 25.9k tokens, inspiration programs fill 46\% of a mutation prompt,
and 20\% of mutation calls yield no tested program.

The OpenEvolve port runs OpenEvolve 0.4.1 at its documented default configuration, with five islands, MAP-Elites on
complexity and diversity, diff-based evolution and four parallel evaluations.
Its calls reach the model through a local endpoint that charges the ledger and refuses calls once the budget is
spent, and every candidate is recorded when OpenEvolve adds it to its database.
It deviates from the defaults in six ways.
It uses one model in place of an ensemble, an output cap of 16,384 tokens in place of 4,096 and a request timeout of
900 seconds.
It sends no top-p because the model rejects it together with a temperature, limits code to 100,000 characters, and
passes the task description as its system message.
Its seed comes from the same repair chain as the ShinkaEvolve port.
AB-MCTS-A runs through Sakana AI's TreeQuest library at its defaults with two expansions at a time.
Expanding the root writes a fresh program, expanding a node refines its program with its feedback, and a node scores
the rank of its program among the valid programs of the run, zero for an invalid one.

\section{Earlier ALE-Bench Batches}
\label{app:earlier}

Table~\ref{tab:earlier} gives the three ALE-Bench batches of the submitted version, which all let the search see 50
public cases and score under our judge.
Every search ran three times, Prog.\ counts programs per problem, and the port runs in ShinkaEvolve's default
configuration (defaults) or with its recommendation and novelty calls on (ours).
The lower block pairs \method with the named search of the same batch, with $\Delta$ and W/T/L computed before
rounding, and official rules replace our judge in the rows that name them.
Only the third batch's contrasts were pre-registered.
An asterisk marks a Holm-adjusted $p$ below 0.05 within the batch, or $p$ below 0.05 for a pre-registered primary or
official-rules contrast.
The dagger and double dagger count zeros on toyota2023summer-final as in Table~\ref{tab:ale}.
In these batches \method leads the port by 162 to 241 points and best-of-N by 136 to 156, while its lead over
sequential refinement is 5 points.
We also ran \method and its variant without early stopping up to 1.8M tokens per problem on the Lite subset, three
runs each in one batch (Table~\ref{tab:budget}).
From 600k to 1.8M tokens \method gains 115 points and the variant 96 points.
At 1.8M the lead of \method over the variant is 70 points without being separated, and most of it comes from the three
Lite problems we used during development.
Both searches keep gaining at a similar rate, so most of the return on a larger budget comes from further refinement,
which the two searches share.

\begin{table}[t]
\centering
\footnotesize
\setlength{\tabcolsep}{3.6pt}
\renewcommand{\arraystretch}{1.1}
\begin{tabular}{@{}lrrrr@{}}
\toprule
Search & Score & Prog. & $\Delta$ & W/T/L \\
\midrule
\multicolumn{5}{@{}l}{\emph{Third batch, pre-registered}} \\
\addlinespace[1pt]
Port, defaults & 1450\sd{85} & 26 & $+162^{*}$ & 32/0/8 \\
\quad official rules & & & $+143^{*}$ & 29/1/10 \\
Port, ours & 1392\sd{29} & 20 & $+220^{*}$ & 35/0/5 \\
Best-of-N & \second 1456\sd{15} & 95 & $+156^{*}$ & 30/0/10 \\
\method{}$^{\ddagger}$ & \best 1612\sd{62} & 39 & & \\
\midrule
\multicolumn{5}{@{}l}{\emph{First batch}} \\
\addlinespace[1pt]
Port, ours & 1427\sd{66} & 19 & $+241$ & 32/0/8 \\
\quad official rules & & & $+202$ & 31/0/9 \\
\method & \best 1668\sd{20} & 40 & & \\
\quad w/o e.s. & \second 1654\sd{16} & 38 & & \\
\midrule
\multicolumn{5}{@{}l}{\emph{Second batch}} \\
\addlinespace[1pt]
Best-of-N & 1491\sd{26} & 95 & $+136$ & 30/0/10 \\
Sequential refinement & 1622\sd{11} & 35 & $+5^{*}$ & 28/0/12 \\
AIDE & 1582\sd{51} & 45 & & \\
\method{}$^{\dagger}$ & 1626\sd{48} & 40 & & \\
\quad w/o e.s. & 1648\sd{39} & 38 & & \\
\quad w/o e.s.\ or notes & \best 1684\sd{23} & 41 & & \\
\quad w/o restarts & \second 1665\sd{37} & 38 & & \\
\quad w/o patience & 1626\sd{25} & 39 & & \\
\bottomrule
\end{tabular}
\vspace{-2pt}
\caption{The three batches of the submitted version. ALE-Bench, 40 problems, Claude Sonnet 4.6, 600k tokens, 50
public cases, our judge unless official rules are named. $\Delta$ and W/T/L compare \method with the search of the row, shading marks the
best two scores per batch, and e.s.\ is early stopping.}
\label{tab:earlier}
\end{table}

\begin{table}[t]
\centering
\footnotesize
\setlength{\tabcolsep}{3.6pt}
\begin{tabular}{@{}lrrrrr@{}}
\toprule
Search & 150k & 300k & 600k & 1.2M & 1.8M \\
\midrule
W/o early stopping & 1395 & 1453 & 1506 & 1547 & 1602 \\
\method & 1407 & 1508 & 1557 & 1602 & 1672 \\
\bottomrule
\end{tabular}
\capgap
\caption{ALE-Bench Lite performance under our judge by token budget, three runs per search in one batch.}
\label{tab:budget}
\end{table}

\section{Results without toyota2023summer-final}
\label{app:toyota}

On the problem toyota2023summer-final some runs of \method score zero because none of their programs passes a public
case within the budget.
Its three runs scored 1727, 1815 and 1884 in the first batch, and all three scored zero in the second.
Two of three scored zero in the third batch, in the fourth batch with five cases and in the fifth batch, while all
three runs of the fourth batch with 50 cases passed public cases.
Best-of-N scored zero in all three runs of the third and fourth batches (Table~\ref{tab:toyotarow}).

We re-ran the 3,779 programs that the first three batches wrote for this problem on its 50 public cases.
Every model call returned a program, 4.8\% of the programs failed to compile and 14.9\% passed every case.
Most of the rest failed every case, 52.5\% with wrong answers, 21.0\% with run-time errors and 3.8\% by exceeding the
time limit.
The model can write a valid program for this problem, and the failure lies in the search.
In all nine runs of \method that scored zero, the first program that the run kept as its best passed no public case.
Programs that pass no case tie, and this program stayed the best.
Every attempt from the best program started from it, and the patience of two turns in Eq.~\ref{eq:patience} stopped
each of these attempts after two turns.
We wrote this cause down before reading the logs of the failing runs.
Sequential refinement never cuts its chain, and 46\% of its programs on this problem were valid in the second batch,
against none of the 198 programs of \method.
Building on the best program only after it passes a public case gave a valid program in two of three runs of the
fourth batch with 50 cases, where plain \method found one in all three runs.
This change does not settle the problem.

Table~\ref{tab:toyota} repeats the contrasts of \method on all 40 problems and on the 39 others, and we report the
subset without a test.
Each value is the mean over problems of the difference between seed-averaged scores, which the reports of the first
three batches round to integers.
Therefore those values on all 40 problems can differ from the main text by a few tenths.
In the fourth batch the lead over the port rises from 140 to 157 points without the problem.
The lead over best-of-N falls from 116 to 103 because best-of-N also scored zero there, and the difference to
independent refinement moves from $-18$ to $+3$.
In the earlier batches every lead over the port rises, except in the first batch, where it falls by about 11 points.
The lead over best-of-N grows in the second batch and shrinks in the third, where best-of-N also scored zero.
The lead over sequential refinement rises from 5 to 52 points in the second batch, while \method keeps the same 28
wins.

\begin{table}[t]
\centering
\footnotesize
\setlength{\tabcolsep}{3.4pt}
\begin{tabular}{@{}lrrrr@{}}
\toprule
Batch & \method & Port, def. & Port, ours & Best-of-N \\
\midrule
First & 1809 & & 1160 & \\
Second & 0 & & & 1052 \\
Third & 608 & 800 & 1086 & 0 \\
Fourth & 625 & 1159 & & 0 \\
Fifth & 573 & & & \\
\bottomrule
\end{tabular}
\capgap
\caption{Scores on toyota2023summer-final, seed-averaged, five public cases in the fourth batch. In the second batch
sequential refinement scores 1848, AIDE 869 and \method without early stopping 1632, and in the first batch \method
without early stopping scores 1761. In the fourth batch sequential refinement scores 1970 and independent refinement
1433, and in the fifth OpenEvolve scores 1445 and AB-MCTS-A 559.}
\label{tab:toyotarow}
\end{table}

\begin{table}[t]
\centering
\footnotesize
\setlength{\tabcolsep}{3.4pt}
\begin{tabular}{@{}lrrrr@{}}
\toprule
& \multicolumn{2}{c}{40 problems} & \multicolumn{2}{c}{Without toyota} \\
\method minus & $\Delta$ & W/T/L & $\Delta$ & W/T/L \\
\midrule
\multicolumn{5}{@{}l}{\emph{Fourth batch, five cases, official rules}} \\
port, defaults & $+139.8$ & 27/1/12 & $+157.1$ & 27/1/11 \\
best-of-N & $+115.6$ & 29/1/10 & $+102.6$ & 28/1/10 \\
seq.\ ref. & $+84.5$ & 28/0/12 & $+121.2$ & 28/0/11 \\
AIDE & $+113.1$ & 26/1/13 & $+128.7$ & 26/1/12 \\
indep.\ ref. & $-17.7$ & 21/1/18 & $+2.6$ & 21/1/17 \\
\midrule
\multicolumn{5}{@{}l}{\emph{Third batch, pre-registered}} \\
port, defaults & $+161.6$ & 32/0/8 & $+170.7$ & 32/0/7 \\
\quad official rules & $+143.2$ & 29/1/10 & $+151.8$ & 29/1/9 \\
port, ours & $+219.7$ & 35/0/5 & $+237.5$ & 35/0/4 \\
\quad official rules & $+236.5$ & 32/0/8 & $+254.8$ & 32/0/7 \\
best-of-N & $+156.0$ & 30/0/10 & $+144.4$ & 29/0/10 \\
\midrule
\multicolumn{5}{@{}l}{\emph{First batch}} \\
port, ours & $+241.1$ & 32/0/8 & $+230.6$ & 31/0/8 \\
\quad official rules & $+202.4$ & 31/0/9 & $+190.9$ & 30/0/9 \\
\midrule
\multicolumn{5}{@{}l}{\emph{Second batch}} \\
best-of-N & $+135.6$ & 30/1/9 & $+166.1$ & 30/1/8 \\
seq.\ ref. & $+4.6$ & 28/0/12 & $+52.1$ & 28/0/11 \\
AIDE & $+44.0$ & 22/1/17 & $+67.5$ & 22/1/16 \\
\bottomrule
\end{tabular}
\capgap
\caption{Contrasts of \method on all 40 ALE-Bench problems and without toyota2023summer-final, our judge unless
official rules are named. The second batch's best-of-N row counts one near-tie as a tie, which the report counts as a
loss.}
\label{tab:toyota}
\end{table}

\section{Notes on Tables 1 to 3}
\label{app:tablenotes}

In Table~\ref{tab:ale}, Score is the mean over problems and runs with the standard deviation over runs.
Every search ran three times, except independent refinement with 50 cases in the fourth batch, which ran five times
for the test of building on the best program in Section~\ref{sec:ana-parts}.
The fourth batch ran every search twice in one launch, once seeing five public cases and once seeing 50, and
the same cases select its submission.
Prog.\ counts programs per problem, in the runs with five cases for the fourth batch.
The port runs in ShinkaEvolve's default configuration, and Seq.\ and Indep.\ abbreviate sequential and
independent in Table~\ref{tab:beyond}.
The right columns give $\Delta$ and W/T/L, the problems won, tied and lost, computed before rounding, and a blank cell
is a contrast we did not test.
In the fourth batch the contrasts of \method with the port and best-of-N were the pre-registered co-primaries,
Holm-adjusted over the pair, and the other five are Holm-adjusted over the five.
In the fifth batch the contrasts with OpenEvolve were the co-primaries, Holm-adjusted over the pair.
Independent refinement minus AB-MCTS-A was a primary of its own and is not adjusted, and the two other contrasts of
\method are Holm-adjusted together with AB-MCTS-A minus OpenEvolve, which leads by 166 points (35/1/4).
Under official rules AB-MCTS-A is 1 point below independent refinement.
The dagger marks a search that scored zero on toyota2023summer-final in all three runs and the double dagger in two of
three runs, counted over the runs with five cases in the fourth batch.
W/o means without in every table.
Appendix~\ref{app:earlier} gives the three batches of the submitted version.

In Table~\ref{tab:lite}, Cost is per problem, and shading marks the best two of our searches in each of our blocks.
The published rows rest on one run for sequential refinement and ALE-Agent and on ten runs for the AIDE agents with
Gemini 3 Flash.
The other published rows report no run count.
Our searches ran three times with Claude Sonnet 4.6 and ten times with Gemini 3 Flash.
Note a marks a cost reported for the run on all 40 problems.
Note b marks a cap of five dollars per problem and run that covers the agent's model calls and the execution of its
programs, with at most 200 steps and 12 hours.
Note c marks a cost that excludes the self-rewriting run that produced the agent, and note e a cap per run, with 36
hours.
Note d marks our Lite-10 average at 600k tokens and list prices without prompt caching.
Note f marks our cap of five dollars per problem and run at list prices of \$0.50 and \$3.00 per million input and
output tokens, which covers model calls only.
Executing the programs of one such run took at most 0.79 CPU-hours, about \$0.08 at an assumed \$0.10 per vCPU-hour.
These runs see five public cases, and their AIDE is our implementation.
Seq.\ ref.\ is sequential refinement, + domain adds domain knowledge, AIDE\textsubscript{human} is a human-built
production agent that \aidetwo reports, and the stronger models are Fable 5 and gpt-5.6-sol.

In Table~\ref{tab:beyond}, every search was frozen after development on ALE-Bench with Claude, and shading marks the
best two scores per column.
CO-Bench and held-out scores are percentages averaged over three runs with Claude Sonnet 4.6, with the standard
deviation over runs.
A CO-Bench score is the percent of the benchmark's reference solution.
A held-out score is the percent of the best score that any of the six searches reached on each test case.
The contrasts in Section~\ref{sec:res-beyond} use the reference fixed for the submitted version, the best of its four
searches per case.
Under it the lead of \method over the port is 15.4 points, and under the reference of the table it is 15.1.
ALE scores are performance points on the 40 problems under our judge, averaged over three runs of GPT-5.2 or Gemini 3
Flash in one batch per model.
We did not run the port with these models (n.r.).

\end{document}